# PANFIS++: A Generalized Approach to Evolving Learning

Mahardhika Pratama

*Abstract* – the concept of evolving intelligent system (EIS) provides an effective avenue for data stream mining because it is capable of coping with two prominent issues: online learning and rapidly changing environments. We note at least three uncharted territories of existing EISs: data uncertainty, temporal system dynamic, redundant data streams. This book chapter aims at delivering a concrete solution of this problem with the algorithmic development of a novel learning algorithm, namely PANFIS++. PANFIS++ is a generalized version of the PANFIS by putting forward three important components: 1) An online active learning scenario is developed to overcome redundant data streams. This module allows to actively select data streams for the training process, thereby expediting execution time and enhancing generalization performance; 2) PANFIS++ is built upon an interval type-2 fuzzy system environment, which incorporates the so-called footprint of uncertainty. This component provides a degree of tolerance for data uncertainty. 3) PANFIS++ is structured under a recurrent network architecture with a self-feedback loop. This is meant to tackle the temporal system dynamic. The efficacy of the PANFIS++ has been numerically validated through numerous real-world and synthetic case studies, where it delivers the highest predictive accuracy while retaining the lowest complexity.

## I. Introduction

Data streams present unique problems which are too costly to be handled by the conventional approach in machine learning. The problem of data streams is frequently encountered in the online time-critical applications, which calls for an efficient algorithm with low computational and storage requirement [1]. Data streams often pertain to the concept drift issue, because it is generated in non-stationary environments. For instance, the manufacturing process runs continuously without any stoppage at a high feed rate. This imposes data streams being generated at a fast sampling rate. Furthermore, the tool condition is monitored at rapidly changing environments, associated with variations of machining parameters, cutter profiles, etc. Tool degradation and changing surface integrity also cause gradual concept drift [2]. To this end, Evolving Intelligent System (EIS) has been proposed and features two prominent characteristics: open structure, online learning [3]. The open structure property allows EIS to start its learning process from scratch with an empty rule base. Its components are self-structured and organized from data streams. This trait provides an answer to non-stationary system behaviors, because it can automatically grow when a change occurs in data distribution. The EIS also characterizes a strict online learning procedure, which can process data streams with high efficiency. A sample is directly discarded once learned, which renders economical computational and memory burdens. This is capable of dealing with a possible infinite nature of data stream and satisfying a life-long learning requirement [3].

In the past decade, the EIS has grown to be a very active research area and drawn considerable research interest of the computational intelligence community. The area of EIS was pioneered by Juang and Lin [4] with SONFIN, although the term "Evolving" has not been formalized until the development of DENFIS [5] and eTS [6]. Since then the area of EIS has attracted various contributions [7] - [15]. Nevertheless, the EIS area deserves in-depth investigation, because three issues, namely data uncertainty, temporal system dynamic, and redundant data streams, are unsolved. Although the concept of metacognitive learning incorporating the what-to-learn component aims at addressing redundant data streams [16]-[20], most of them are designed for classification problems, while a regression problem is an open issue. They make use of the so-called hinge error function to approximate data stream contribution. It is at risk of the bias-and-variance dilemma because system error can be high in the case of overfitting. Some attempt has been devoted to actualise the evolving concept in the recurrent network structure and the interval-type 2 fuzzy system [21]-[25]. We argue however that these works are over-dependent on the KM iterative procedure, incurring costly computational burden. These works are built upon a distance-based clustering method, which is prone to outliers. In addition, these methods do not possess the self-regulatory mechanism that controls data streams to train the model. As a result, all incoming data streams are subject to the training process. It slows down the training process and even brings down the generalization capability because redundant data streams are perhaps utilized to train the model.

This paper presents a novel evolving learning algorithm, called Parsimonious Network Based on Fuzzy Inference System++ (PANFIS++). PANFIS++ presents an extended version of PANFIS [26], which is not only capable of mining data streams efficiently but also selecting data streams to be learned on the fly. The salient components of PANFIS++ are elaborated as follows: 1) The online active learning for regression problem, namely ESEM, is proposed in the PANFIS++ and is capable of ruling out inconsequential samples, thereby improving the generalization's performance and expediting the training process; 2) PANFIS++ is constructed under a recurrent network structure, which features a local recurrent loop. The recurrent network architecture aims at capturing temporal system dynamics and alleviating the need for time-delayed input attributes. Although the local recurrent connection has been put into perspective in [21], [22], the novelty of our work can be found in the rule layer of the PANFIS++, which features the interval type-2 multivariate Gaussian function with the interval-valued centroids; 3) PANFIS++ realises a generalized interval type-2 Takagi-Sugeno-Kang (TSK) fuzzy rule. That is, the first-order polynomial is embedded in the rule consequent, while the multivariate Gaussian function with uncertain Centroids is incorporated in the rule premise; 4) As with other EISs, the PANFIS++ is capable of automatically generating its fuzzy rules in the single-pass learning mode using the Generalized Type-2 Datum Significance (GT2DS) method. The GT2DS is a generalized version of the T2DS method [7], which is designed under a strict condition of uniformly distributed training data. It adopts a strategy of [27], [28], which estimates a complex probability density function with the Gaussian Mixture Model (GMM). It is worth noting that [27], [28] are all designed for the type-1 fuzzy system; 5) PANFIS++ integrates the rule pruning scenario, which is capable of discarding an outdated rule – no longer relevant to current learning context and an inconsequential rule – plays little role during its lifespan. This is carried out using the Type-2 Relative Mutual Information (T2RMI) method; 6) PANFIS++ is equipped with a rule recall scenario, which makes possible to reactivate a previously pruned rule. This scenario is needed to deal with the cyclic concept drift, which presents a previously learned concept. 7) The parameter learning scenario is undertaken using a combination of the Fuzzily Weighted Generalized Recursive Least Square (FWGRLS) method and the Zero Error Density Maximization (ZEDM) principle to adapt other free parameters. The efficacy of the PANFIS++ has been numerically validated with various real-world and synthetic data streams. PANFIS++ has been benchmarked with prominent algorithms, recently published in the literature where PANFIS++ demonstrates more encouraging performance in attaining a balance between accuracy and simplicity than its counterparts.

The remainder of this paper is organized as follows: Section 2 outlines the network architecture of the PANFIS++; Section 3 discusses the rule base management of the PANFIS++; Section 4 details our numerical study and comparisons with state-of-the-art algorithms. Concluding remarks are drawn in the last section of this paper.

## II. Network Architecture of PANFIS++

PANFIS++ is built upon a recurrent network structure using a local recurrent connection, which is meant to tackle the temporal system dynamic and to reduce the demand of time-delayed input variables. Furthermore, the hidden layer of the PANFIS++ is composed of the interval-valued multivariate Gaussian function. The multivariate Gaussian function evolves a flexible ellipsoidal cluster in any direction, which is compatible to cover irregular data, which are not distributed in the main axes. This trait is capable of decreasing the fuzzy rule demand. Furthermore, the multivariate Gaussian function features the scale-invariant property, which can deal with different ranges of input attributes. Another appealing property is its non-diagonal covariance matrix, which retains the interrelation among input variables. The interrelation among input variables vanishes under a conventional fuzzy rule with the *t-norm* operator. Unlike the standard TSK fuzzy rule, the PANFIS++ extends the degree of freedom in the rule consequent with the Chebyshev function. It is well-known that the zero or first-order polynomial do not fully exploit the local mapping capabilities. This drawback is usually addressed by putting forward a higher order polynomial in the rule consequent but this strategy also increases the variance of the model, which increases the likelihood of overfitting. The Chebyshev function is chosen to map the original input space here because it outperforms the mapping capability of other polynomial function. The final output of the PANFIS++ is written as follows:

$$y_o = fo^6 = \frac{(1-q_o)\overline{\psi}_i \beta_i}{\sum_{i=1}^{R} \underline{\psi}_i} + \frac{q_o \underline{\psi}_i \beta_i}{\sum_{i=1}^{R} \overline{\psi}_i} \quad (1)$$

where $q$ stands for the design coefficient and $R$ stands for the number of fuzzy rules. The $q$-design factor is used to perform a type reduction mechanism and is dynamically adapted to be well-suited to the current training context. It controls the proportion of upper and lower fuzzy variables. $\tilde{\psi} = [\underline{\psi}, \overline{\psi}]$ is an interval-valued temporal firing strength, which results from a local recurrent connection. The interval-valued temporal firing strength is expressed as follows:

$$\overline{\psi}_i = \lambda_i \overline{R}_i + (1-\lambda_i)\overline{R}_i, \quad \underline{\psi}_i = \lambda_i \underline{R}_i + (1-\lambda_i)\underline{R}_i \quad (2)$$

where $\tilde{R}_i = [\underline{R}_i, \overline{R}_i]$ denotes an interval-valued spatial firing strength, which is resulted from the interval-valued multivariable Gaussian function. It is worth noting that the uncertainty elements in terms of an interval is embedded in the Centroid of Gaussian function. In other words, the centroid of the Gaussian function is not a crisp vector rather an interval-valued vector. The interval-valued multivariable Gaussian function is formalised as follows:

$$\underline{R}_i = \exp(-(X-\underline{C}_i)\Sigma_i^{-1}(X-\underline{C}_i)^T), \quad \overline{R}_i = \exp(-(X-\overline{C}_i)\Sigma_i^{-1}(X-\overline{C}_i)^T) \quad (3)$$

where $\tilde{C}_i = [\underline{C}_i, \overline{C}_i] \in \Re^{1 \times p}$ is an interval-valued Centroid of the *i-th* rule, $\Sigma_i^{-1} \in \Re^{p \times p}$ is a non-diagonal covariance matrix and $p$ is the number of input dimension. Nevertheless, (3) should not be implemented directly, because $\overline{R}_i > \underline{R}_i$ does not always hold. This problem is resolved by transforming a high dimensional cluster to its one-dimensional equivalent to satisfy the basic principle of interval arithmetic. In other words, a fuzzy set representation of a high-dimensional cluster needs to be obtained before performing the fuzzy inference scheme. This is done by simply enumerating radii of the Gaussian fuzzy set from a high dimensional ellipsoidal cluster. The radii of the fuzzy set are formulated as a distance between the centres to the cutting points of the ellipsoidal cluster [26] as follows:

$$\sigma_i = \frac{(\overline{r}_i + \underline{r}_i)}{2} \sqrt{\Sigma_{i,j,j}^{-1}} \tag{4}$$

where $\Sigma_{i,j,j}^{-1}$ is diagonal elements of the inverse covariance matrix of the *i-th* rule. It is worth noting although this strategy is relatively simple to use and fast, it is rather inaccurate provided that an ellipsoidal cluster is rotated around 45 degrees. The centre of the Gaussian fuzzy set adopts a centre of a multidimensional ellipsoidal cluster without any modification. After obtaining the fuzzy set representation of the Gaussian function, the firing strength is calculated as follows:

$$\tilde{\mu}_{i,j} = \exp\left(-\frac{(x_j - c_{i,j})^2}{\sigma_{i,j}^2}\right) = N(x_j; \tilde{c}_{i,j}, \sigma_{i,j}), \tilde{c}_{i,j} = [\underline{c}_{i,j}, \overline{c}_{i,j}] \tag{5}$$

$$\overline{\mu}_{i,j} = \begin{cases} N(x_j; \overline{c}_{i,j}, \sigma_{i,j}), x_j < \overline{c}_{i,j} \\ 1, \underline{c}_{i,j} \leq x_j \leq \overline{c}_{i,j} \\ N(x_j; \underline{c}_{i,j}, \sigma_{i,j}), x_j > \underline{c}_{i,j} \end{cases} \tag{6}$$

$$\underline{\mu}_{i,j} = \begin{cases} N(x_j; \overline{c}_{i,j}, \sigma_{i,j}), x_j \leq \frac{(\underline{c}_{i,j} + \overline{c}_{i,j})}{2} \\ N(x_j; \underline{c}_{i,j}, \sigma_{i,j}), x_j \geq \frac{(\underline{c}_{i,j} + \overline{c}_{i,j})}{2} \end{cases} \tag{7}$$

The membership degree of all dimensions are combined with a conjunction operator, namely the product *t-norm* operator as follows:

$$\underline{R}_i = \prod_{j=1}^{p} \underline{\mu}_{i,j}, \overline{R}_i = \prod_{j=1}^{p} \overline{\mu}_{i,j} \tag{8}$$

The firing strength is widely used to measure a compatibility of existing rules in the rule growing scenario. The flaw of this approach is, however, outliers, because it is functionally equivalent to the distance-based clustering approach. The product *t*-norm operator suffers from the curse of dimensionality problem, where (8) diminishes proportionally to the number of input features.

The rule consequent of the PANFIS++ is built upon a nonlinear mapping through up to the second order of the Chebyshev polynomial $\beta_i = x_e W_i$, where $x_e \in \Re^{1\times(2p+1)}$ denotes an extended input vector based on the Chebyshev polynomial, while $W_i \in \Re^{(2p+1)\times 1}$ stands for the weight vector of the *i-th* rule. The Chebyshev polynomial is defined as follows:

$$A_{n+1} = 2x_j A_n(x_j) - A_{n-1}(x_j) \tag{10}$$

Suppose that we deal with 2D-case $X \in \Re^{1\times 2}$, the extended input vector $x_e$ is formulated as $x_e = [1, A_1(x_1), A_2(x_1), A_1(x_2), A_2(x_2)]$. The term 1 is inserted in the extended input vector as an intercept to provide flexibility and to avoid an untypical gradient case.

## III. Learning Policy of PANFIS++

This section elaborates the rule base management of the PANFIS++: Section III.A outlines the online active learning procedure of the PANFIS++, Section III.B discusses the rule growing scenario of the PANFIS++, Section III.C covers the rule pruning scenario of the PANFIS++, Section III.D describes the rule recall scenario of the PANFIS++, Section III.D details the parameter learning scenario of the PANFIS++. An overview of overall training procedures of PANFIS++ is depicted in Algorithm 1.

*III.A An Online Active Learning Scenario*

The online active learning scenario is capable of expediting the training process because it can reduce the number of training samples to be seen. It also prevents redundant samples to be learned. This contributes to the significant improvement of generalization power. Although the active learning concept has been well-established in the literature, most of them work in the pool-based approach, which assumes all data are available in the pool for an iterative selection procedure [29]. Recently, some initiatives were devoted to proposing an online active learning scenarios to mine data streams [30], [31]. An open research is, however, the regression case, where a target variable is a continuous function. The regression problem requires more in-depth investigation than the classification problem because the sample contribution cannot be examined with respect to its position to a decision surface. One technique has been put forward in [31] but it is still based on the hinge loss function, which is sensitive to the system's error. As a matter of fact, the system error does not necessarily reflect a true system condition, because a high variance case also results in a high system error.

The PANFIS++ utilizes the ESEM to steer the online active learning strategy. The ESEM relies on the entropy of neighborhood probability, which studies uncertainties of existing structure with respect to an incoming sample. The ESEM is a generalization of SEM [32] which forms an incremental version of the SEM. The probability of a data stream to sit within influence zone of existing rules – the neighborhood probability - is formalized as follows:

$$P(X_n \in N_i) = \frac{\sum_{n=1}^{N_i} M(X_N, x_n)}{\sum_{i=1}^{R} \sum_{n=1}^{N_i} M(X_N, x_n)} \quad (11)$$

where $X_N, x_n, M(\ ), N_i$ are respectively the newly seen sample, the *n-th* sample of *i-th* rule, a similarity measure, and the number of populations of *i-th* cluster. (11) is intractable for the online scenario, because it revisits all samples seen thus far. It is obvious that $\sum_{n=1}^{N_i} M(X_N, x_n)$ is akin to the local density since it looks at spatial proximities between a sample and all populations of the *i-th* rule. It also signifies a robust approach against uncertainty, since it is resulted from all collected samples. A recursive expression of $\sum_{n=1}^{N_i} M(X_N, x_n)$ can be defined:

$$\frac{\sum_{n=1}^{N_i} M(X_N, x_n)}{N_i} = \frac{\sum_{j=1}^{p}(N_i-1)x_j^N - 2\sum_{j=1}^{P} x_j^N \vartheta_j^{N_i} + \vartheta_{N_i}}{(N_i-1)u} \quad (12)$$

$$\vartheta_j^{N_i} = \vartheta_j^{N_i-1} + x_j^{N_i-1}, \vartheta_{N_i} = \vartheta_{N_i-1} + \sum_{j=1}^{u} x_{N_i-1,j}^2$$

The final formula of the ESEM is exhibited as an entropy of the neighborhood probability:

$$H(N|X_N) = -\sum_{i=1}^{R} P(X_N \in N_i) \log P(X_N \in N_i) \quad (13)$$

An attention should be paid to a sample incurring a high level of uncertainty, whereas a sample with low uncertainty is negligible to the training process. Such sample can be ignored without loss of generalization, thus speeding up the training process. Hence, the condition to accept a sample for the training process is shown as follows:

$$H(N|X_N) \geq \delta_1 \quad (14)$$

where $\delta_1$ is a predefined threshold. In non-stationary environments, $\delta_1$ should not be fixed during the training process rather be adjustable to adapt to system's behaviour. This strategy is confirmed with the fact that shifts in the data distribution entail different levels of the threshold. A heuristic approach is adopted to fine-tune the threshold where the threshold should augment when a new training sample is accepted $\delta_1^{N+1} = \delta_1^N(1 + s)$. This aims to reduce possibility of next training samples to be used in the training process. It should on the other hand decline when a new data point is admitted for model updates $\delta_1^{N+1} = \delta_1^N(1 - s)$. The goal is to boost the chance of next samples to be next training patterns. In other words, this adjustment aims to achieve tradeoff between complexity and accuracy. We set *s* as its default value [33] *s=0.01*.

*III.B Rule Growing Mechanism*

PANFIS++ is capable of automatically evolving its fuzzy rules on demands in accordance with inherent data distributions using the GT2DQ method. The DQ method relies on the idea of statistical contribution estimation of a data stream, which includes the possible future contribution of a data point. The original version of DQ method [7] is derived with the *p-fold* numerical integration, which is only suitable for a small input dimension. Moreover, it assumes that data are uniformly distributed. This fact renders the DQ method less accurate when data samples are sparsely distributed. The GT2DQ method is proposed here to overcome these drawbacks of T2DQ. It is inspired by the works of [27], [28] taking advantage of the GMM as a probability density function to deal with a complex and irregular data distribution. In light of the rule significance definition [7], the significance of interval-valued multivariate Gaussian function is defined as the *$L_u$-norm* of the error function weighted by the input density function. This leads to the final expression of the GT2DQ method as follows:

$$E_i = \|\beta_i\|(1-q)\left\{(2\pi/u)^{p/2} det(\Sigma_i)^{-1/2} \overline{N}_i \gamma^T\right\}^{\frac{1}{u}} + \|\beta_i\|q\left\{(2\pi/u)^{p/2} det(\Sigma_i)^{-1/2} \underline{N}_i \gamma^T\right\}^{\frac{1}{u}}$$

$$\overline{N}_i = \left[N(\overline{c}_i - v_1; 0, \Sigma_i^{-1}/u + \Sigma_1), (\overline{c}_i - v_2; 0, \Sigma_i^{-1}/u + \Sigma_2), \ldots, (\overline{c}_i - v_m; 0, \Sigma_i^{-1}/u + \Sigma_m), \ldots, (\overline{c}_i - v_M; 0, \Sigma_i^{-1}/u + \Sigma_M)\right],$$
$$\underline{N}_i = \left[N(\underline{c}_i - v_1; 0, \Sigma_i^{-1}/u + \Sigma_1), (\underline{c}_i - v_2; 0, \Sigma_i^{-1}/u + \Sigma_2), \ldots, (\underline{c}_i - v_m; 0, \Sigma_i^{-1}/u + \Sigma_m), \ldots, (\underline{c}_i - v_M; 0, \Sigma_i^{-1}/u + \Sigma_M)\right] \quad (15)$$

where $v_m, \Sigma_m, \alpha_m$ are respectively the parameters of GMM $\sum_{o=1}^{m} \alpha_o N(x; v_o, \Sigma_o)$ and the mixing coefficients $\sum_{o=1}^{M} \alpha_o = 1, \alpha_o > 0$. It is worth mentioning that the GMM parameters can be elicited using pre-recorded samples $N_{history}$. The use of pre-recorded samples still makes sense in practise especially in the big data era, where access to pre-recorded samples is not difficult to be obtained. The number of pre-recorded samples in addition is significantly smaller than the number of training samples $N_{history} \ll N$. We find also that $N_{history}$ is not problem-specific and is fixed at 30 for all simulations in this paper.

The rule growing procedure is undertaken first of all by creating a hypothetical rule $\overline{C}_{R+1}, \underline{C}_{R+1}, \Sigma_i^{-1}$ from a newly seen sample as follows:

$$\tilde{C}_{R+1} = X_N \pm \Delta, \Sigma_{R+1} = \frac{\max((C_i - C_{i-1}),(C_i - C_{i+1}))}{\sqrt{\ln(1/\epsilon)}} I \quad (16)$$

where $\epsilon$ is a $\epsilon$ factor set in respect to the desired completeness degree. We here follow the commonly used setting in the literature $\epsilon = 0.5$. $\Delta$ is an uncertainty threshold, which governs the footprint of uncertainty of the interval-valued multivariate Gaussian function. We set $\Delta = 0.1$ for all simulations in this paper for simplicity. The initialization of the covariance matrix (16) has been proven mathematically to meet the $\epsilon$ – completeness condition [34]. The hypothetical rule is added to be a new rule when it offers a substantial statistical contribution over existing rules. A rule growing condition is formalised as follows:

$$E_{R+1} \geq \max_{i=1,\ldots,R} E_i \qquad (17)$$

This rule growing condition differs from its predecessors [7], [27], [28], which are reliant on a user-defined threshold. The threshold is often problem-dependent and entails substantial expert knowledge to arrive at a correct value. (17) is a plausible condition for a rule generation procedure, because the statistical contribution of the hypothetical rule exceeds existing rule. Furthermore, the GT2DQ is relatively robust against outliers because the statistical contribution is derived from the whole region defined by a complex input density function set as the GMM.

The GT2DQ method cannot be standalone in the rule growing procedure of the PANFIS++, because it yet analyses the position of a sample in the input space. This is necessary to illustrate the significance of a data sample to a current network structure and to arrive at a proper input space clustering. We to remedy this problem insert a compatibility measure, which delves a spatial proximity of a sample to current cluster prototypes. The compatibility measure is realized using the interval-valued spatial firing strength as depicted in(3) executed with the q-design factor to produce its crisp values as follows:

$$FS \leq \delta_2, FS = q\underline{R_i} + (1-q)\overline{R}_i \qquad (18)$$

where $\delta_2$ is a threshold, which determines a minimum conflict level and is assigned as the critical value of the chi-square distributed $\chi^2$ with $p$ degree of freedom and a significance level of $\alpha$. $\delta_2$ is then allocated as $\delta_2 = \exp(-\chi^2)$ with the significance level $\alpha = 5\%$. This condition assures that a new data point is sufficiently remote from the zone of influence of existing rules. Because PANFIS++ makes use of the multivariate Gaussian function, the chi-square distribution can be referred as a basis of a threshold selection, because it statistically formulates a condition when a sample is out of coverage area. If (17) and (18) are complied, a hypothetical rule is introduced as a new rule and its rule premise is specified as (16), while its rule consequence is set as follows:

$$W_{R+1} = W_{win}, \Psi_{R+1} = \omega I \qquad (19)$$

where $\omega = 10^5$ is a positive large value and heuristically fixed at $10^5$. This setting has been analytically proven to be a close approximation of a batched learning scheme. Furthermore, a new rule consequent is akin to that of the winning rule, because the winning rule should represent similar output trend of the winning rule. There are several ways to choose the winning rule: the distance-based method, the Bayesian concept [35]. The Bayesian concept is applied in the PANFIS++ because of its prior probability. A winning rule is selected as the one with the highest posterior probability. It is not recounted here and interested reader is advised to go to [35] for further details of the Bayesian winning rule selection.

If (17) and (18) are not satisfied, a data sample does not feature sufficient statistical contribution to trigger the rule growing scenario. This sample should be absorbed to refine the position of the fuzzy rule in the data space. Specifically, the rule antecedent is fine-tuned using this sample as follows:

$$\tilde{C}_{win}^{N} = \frac{N_{win}^{N-1}}{N_{win}^{N-1}+1}\tilde{C}_{win}^{N-1} + \frac{(X_N - \tilde{C}_{win}^{N-1})}{N_{win}^{N-1}+1}, \tilde{C}_{win} = [\underline{C}_{win}, \overline{C}_{win}] \quad (20)$$

$$\Sigma_{win}(N)^{-1} = \frac{\Sigma_{win}(N-1)^{-1}}{1-\alpha} + \frac{\alpha}{1-\alpha}\frac{(\Sigma_{win}(N-1)^{-1}(X_N - \hat{C}_{win}^{N-1}))(\Sigma_{win}(N-1)^{-1}(X_N - \hat{C}_{win}^{N-1}))^T}{1+\alpha(X_N - \hat{C}_{win}^{N-1})\Sigma_{win}(old)^{-1}(X_N - \hat{C}_{win}^{N-1})^T} \quad (21)$$

$$N_{win}^{N} = N_{win}^{N-1} + 1 \quad (22)$$

where $\alpha = 1/(N_{win}+1)$ and $\hat{C}_{win} = (\overline{C}_{win} + \underline{C}_{win})/2$. The midpoint of upper and lower bounds of the interval-valued cluster prototypes to adjust the crisp covariance matrix. Moreover, a direct adaptation of the inverse covariance matrix is put forward where no reinversion step is required after performing any adaptation phase. The reinversion step incurs computationally prohibitive cost notably in the high input dimension case. It tends to be unstable in the light of reinversion step when a covariance matrix is ill-defined.

*III.C Parameter Learning Scenario of PANFIS++*

The PANFIS++ puts forward a synergy between the zero-error density maximization method (ZEDM) and the FWGRLS method to perform the parameter learning scenario of the PANFIS++. The ZEDM is used to adjust the $q$-design factor of PANFIS++ and constitutes a generalized version of the gradient descent method [35]. It substitutes the Mean Square Error (MSE) with the error entropy as the cost function because it turns out to be more thorough to capture the high order statistical behavior than the MSE. Minimizing the error entropy is equivalent to reducing the distance between the probability distribution of the target function and the predictive output. Since the model of the error entropy is extremely difficult to formalize with the first principal, it is approximated with the Parzen Window technique here as follows:

$$\hat{f}(0) = \frac{1}{N\tau\sqrt{2\pi}}\sum_{n=1}^{N}\exp(-\frac{e_n^2}{2\tau^2}) \quad (23)$$

where $\tau$ is a smoothing parameter, simply set as 1 in our case and $e_n$ is the system error, while $N$ is the number of training samples seen by far. The optimisation procedure follows the stochastic gradient descent approach where a gradient computation and adaptation are carried out per sample. This is defined as follows:

$$q_o^N = q_o^{N-1} - \eta_q \frac{1}{N\sqrt{2\pi}} \frac{\partial E}{\partial q_0} \sum_{n=1}^{N} \exp(-\frac{e_n^2}{2}) \quad (24)$$

$$\lambda_i^N = \lambda_i^{N-1} - \eta_\lambda \frac{1}{N\sqrt{2\pi}} \frac{\partial E}{\partial \lambda_i} \sum_{n=1}^{N} \exp(-\frac{e_n^2}{2}) \quad (25)$$

where $\eta_q$ denotes a learning rate. $\frac{\partial E}{\partial q}$ is the gradient of the squared error $\frac{(y-t)^2}{2}$ in respect to the $q$ design coeffient as follows:

$$\frac{\partial E}{\partial q} = e_n \left( \frac{\sum_{i=1}^{R} \overline{\Psi}_i \beta_i}{\sum_{i=1}^{R} \underline{\Psi}_i} - \frac{\sum_{i=1}^{R} \underline{\Psi}_i \beta_i}{\sum_{i=1}^{R} \overline{\Psi}_i} \right) \quad (26)$$

$$\frac{\partial E}{\partial \lambda_i} = e_n \left( \left( (\overline{R}_i - \overline{\Psi}_i) \frac{(1-q)\beta_i}{\sum_{i=1}^{R} \underline{\Psi}_i} \right) + \left( (\underline{R}_i - \underline{\Psi}_i) \frac{q\beta_i}{\sum_{i=1}^{R} \overline{\Psi}_i} \right) \right) \quad (27)$$

It is observed that $\sum_{n=1}^{N} \exp(\frac{-e_n^2}{2})$ in (24), (25) requires revisiting all preceding samples, which do not fit the online learning scenario. Its recursive version is to remedy this bottleneck developed as $A_N = A_{N-1} + \exp(\frac{-e_n^2}{2})$. The learning rate plays vital role to the success of the adaptation process. A too small learning rate leads to slow convergence, whereas a too large learning rate causes instability. Hence, it should be adjusted in respect to the real system behaviour to expedite the convergence as follows:

$$\eta = \begin{cases} \delta_3 \eta, \hat{f}_N(0) \geq \hat{f}_{N-1}(0) \\ \delta_4 \eta, \hat{f}_N(0) < \hat{f}_{N-1}(0) \end{cases} \quad (28)$$

where $\delta_3 \in (1,1.5], \delta_4 \in (0.5,1]$ stand for adjustment factors, which navigate the dynamic of learning rates. We set these parameters as $\delta_3 = 1.1, \delta_4 = 0.9$. The adjustment should be carried out with care because the learning rate should hover around a stable region where convergence is guaranteed. The stable range can be derived using the Lyapunov stability concept and results in the condition $0 < \eta < \frac{2N\sqrt{2\pi}}{(P_0)^2 A_N}$. The proof is left to reader here, because its derivation is well-known in the literature [36].

The FWGRLS method is utilized to adapt the rule consequent of the PANFIS++. The FWGRLS presents a local learning version of the GRLS method [37] which incorporates the weight decay term to improve model's generalization and robustness of the adaptation process. The weight decay term prevents the rule consequent to be too large, which is the main cause of overfitting. In addition, it maintains the weight vector to be in a small and bounded range.

Therefore, an inconsequential rule will be in very small values, which make them easy to detect and to prune. The local learning scenario is compatible with the evolving learning framework because it adapts per rule separately. Hence, convergence and stability of other rules are not affected by the tuning of a rule.

IV. Proof of Concepts

This section discusses the proof of concepts of the PANFIS++. The PANFIS++ was numerically validated through two real-world case studies: valuation of residential premise price and tool wear prediction of a high-speed milling process. The PANFIS++ was compared against 10 prominent learning algorithms: eT2Class [9], Simp_eTS [3], eTS [6], BARTFIS [38], PANFIS [26], GENEFIS [11], DFNN [39], GDFNN [34], ANFIS [40] and benchmarked algorithms were compared against five evaluation criteria: predictive accuracy, fuzzy rule, input attribute, runtime, training sample, and network parameters.

*IV.A   An Appraisal of Residential Premise Price*

This numerical study aims to numerically validate the efficacy of the PANFIS+ to perform a valuation of residential premise price (Courtesy of Dr. Lughofer, Linz). As prevalent characteristic of real-world financial data, this problem is highly volatile and varies overtime, because it is highly influenced by various external parameters: infrastructures, economic activities in the surrounding area, etc. Data were collected from one of the large cities in Poland with 980 K populations. 50 K historical data were recorded during an 11 years period from 1998 until 2008. 5 input features were extracted from property market expert domain knowledge: usable area of premises, the age of a building, number of rooms in a flat, floor on which a flat is located, and distance from the city center [41]. The experimental procedure follows the predictive hold-out process, where 5 consecutive-years data formed the training set, while the testing set was compiled with the subsequent one-year data. The experiment was conducted in the absence of time-lagged input attributes. Table 2 sums up averaged numerical results across the whole periodic hold-out process.

From Table 2, it is observed that the PANFIS++ produced the most encouraging performance in reaching tradeoff between accuracy and complexity. It underwent the fastest execution time because it is fitted by the online active learning strategy. This learning mechanism is capable of suppressing the number of training samples for model updates where in total only 5% of total training data were utilized. In terms of accuracy, the PANFIS++ delivered accurate predictive accuracy in which it came second after the GDFNN. It is worth noting the GDFNN is not suitable for online real-time deployment because it incurs prohibitive complexity as a result of revisiting already seen samples.

Table 2. Appraisal of Residential Premise Price

| ALGORITHMS | Type | MSE | RULE | RUNTIME | SAMPLE | PARAMETERS |
|---|---|---|---|---|---|---|
| PANFIS++ | S-2-R | 0.01±0.002 | 2.2±0.4 | **0.1±0.03** | 98.2 | 119.2 |
| GENEFIS | S-1-F | 0.0119±0.0011 | 5.33±3.14 | 0.3±0.2 | 2417.7 | 78.18 |
| PANFIS | S-1-F | 0.0159±0.0051 | 6.33±4.14 | 0.25±0.04 | 2417.7 | 227.9 |
| eTS | S-1-F | 0.0497±0.0682 | 11±1.89 | 0.22±0.02 | 2417.7 | 126 |
| Simp_eTS | S-1-F | 0.03±0.009 | 3.3±1 | 0.27±0.05 | 2417.7 | **41.7** |
| BARTFIS | S-1-F | 0.0584±0.0223 | 10.2±0.1 | 0.21±0.09 | 2417.7 | 184 |
| ANFIS | S-1-F | 0.0250±0.0235 | 32 | 376.5±0.1 | 2417.7 | 50512 |
| DFNN | B-1-F | 0.004±0.07 | 33.1±0.01 | 543.2±1.1 | 2417.7 | 2869.1 |
| GDFNN | B-1-F | **0.001±0.06** | 9.67±1 | 67.8±15.5 | 2417.7 | 2529 |
| FAOS PFNN | B-1-F | 0.04±0.03 | 11.5±0.55 | 0.33±0.05 | 2417.7 | 126 |
| eT2Class | S-1-F | 0.03±0.009 | **2** | 4.4±1.3 | 2417.7 | 72 |

S: Sequential, B: Batch, 1: type-1, 2: Type 2, R: Recurrent, F: Feed-forward

Table 3. Tool condition monitoring problem of a complex manufacturing process

| Model | Type | RMSE | Rule | Input | Runtime | Sample | Parameters |
|---|---|---|---|---|---|---|---|
| PANFIS++ | S-2-R | **0.04±0.01** | **2** | 12 | **0.7±0.11** | **516.5** | 432 |
| eT2Class | S-2-R | 0.05±0.01 | 5.9±01 | 12 | 1.1±0.7 | 572 | 2065 |
| PANFIS | S-1-F | 0.045±0.01 | **2** | 12 | 0.98±0.1 | 572 | 636.7 |
| GENEFIS | S-1-F | 0.0418±0.008 | 4.1±0.9 | 11.1±0.3 | 1.12±0.1 | 572 | 636.7 |
| Simp_eTS | S-1-F | 0.26±0.08 | 5 | 12 | 1.4±0.3 | 572 | **137** |
| eTS | S-1-F | 0.046±0.01 | 5.1±0.3 | 12 | 1.3±0.02 | 572 | 139.5 |
| BARTFIS | S-1-F | 0.0632±0.01 | 20.6±4. | 12 | 1.43±0.0 | 572 | 762.2 |
| FAOSPFNN | B-1-F | 0.27±0.003 | 12.7±0.7 | 12 | 1.91±0.1 | 572 | 902.2 |
| DFNN | B-1-F | 0.12±0.1 | 101.9±2 | 12 | 24.3±2.4 | 572 | 4444.2 |
| GDFNN | B-1-F | 0.05±0.06 | 5.3±1.15 | 12 | 7.25±1.5 | 572 | 709.2 |
| ANFIS | B-1-F | 0.05±0.01 | 11 | 12 | 183.05±5 | 572 | 979 |

S: Sequential, B: Batch, 1: type-1, 2: Type 2, R: Recurrent, F: Feed-forward

*IV.B Tool Wear Prediction of A High-Speed Milling Process*

This case study is a tool condition monitoring problem of a high-speed milling process, namely the ball nose end milling process (Courtesy of Dr. X. Li, Singapore). It aims to predict the tool wear of the ball nose end milling cutter during the dry machining of hardened tool steel with a hardness of 52-54 HRC. The tool condition monitoring problem of a high-speed milling problem, in general, remains a very complex issue for both academic and industrial practitioners because of the use of multi-point cutting tools at high speed, varying machining parameters, inconsistency, and variability of cutter geometry/dimensions. The experiment moreover took place in non-stationary fashions because cutter degradation often leads to the gradual concept drift. The machining process applied varying spindle rates which causes variations of the data trends. The CNC milling process (Rőders Tech RFM760) was used and raw data were collected using the 7-channels DAQ which correspond to the cutting and force signals in three different cutting axes (X, Y, Z) in addition to the AE signal. The predictive task was carried out using 12 time-domain features extracted from the force signal [1]. It is well-known that the force signal conveys the most informative indicators of the tool wear in respect to the other two signals. A total of 630 data points were obtained from a real manufacturing process and the 10-fold cross validation (CV) scheme was followed as the experimental procedure. Table 3 tabulates an average of numerical results across the 10-fold CV procedure.

From Table 3, it is obvious that the PANFIS++ attained the highest predictive accuracy while retaining the lowest complexity. The PANFIS++ outperformed its counterparts in three evaluation criteria, namely RMSE, the number of rules, the number of samples and runtime. Note that the characteristic of the tool condition monitoring problem is highly dynamic and this result confirms adaptive and evolving traits of the PANFIS++. The recurrent network structure increases adaptivity of a model because it memorizes previous learning actions, combined with the most recent observation.

## V. Conclusion

A generalized version of PANFIS, namely PANFIS++, is proposed in this paper. PANFIS++ introduces a new paradigm of the EIS by providing concrete solutions against three prominent issues: temporal system dynamic, data uncertainty and absence of system order. The PANFIS++ puts forward three novel learning components: 1) A new online active learning strategy for regression problems is put forward where it is capable of expediting the training process while improving model's generalization; 2) The PANFIS++ is constructed under the type-2 fuzzy system environment which incorporates the so-called footprint of uncertainty. This component provides an effective avenue in dealing with the data uncertainty issue; 3) The PANFIS++ makes use a self-feedback loop at the rule layer to cope with the temporal system dynamic and the absence of system order. The self-feedback loop functions as an internal memory component and generates the spatiotemporal firing strength. The efficacy of the PANFIS++ was numerically validated using two real-world data streams containing non-stationary components. It was compared against prominent algorithms in the literature and it is shown that the PANFIS++ attained state-of-the-art performance.